\pdfoutput=1
\documentclass[11pt]{article}

\usepackage{ACL2023}
\usepackage{times}
\usepackage{latexsym}

\usepackage[T1]{fontenc}
\usepackage[utf8]{inputenc}

\usepackage{microtype}

\usepackage{inconsolata}

\usepackage{color}
\usepackage{colortbl}
\usepackage{graphicx}
\usepackage{multirow}
\usepackage{makecell}
\usepackage{float}
\usepackage{subfig}
\usepackage{bm}
\usepackage{upgreek}
\usepackage{booktabs}
\usepackage{amsfonts,amssymb}
\usepackage{amsmath}
\usepackage{bbding}
\usepackage{array}

\usepackage[most]{tcolorbox}
\usepackage{url}

\usepackage{fontawesome}

\makeatletter
\newcommand{\thickhline}{%
    \noalign {\ifnum 0=`}\fi \hrule height 1pt
    \futurelet \reserved@a \@xhline
}
\newcolumntype{"}{@{\hskip\tabcolsep\vrule width 1pt\hskip\tabcolsep}}
\makeatother
\usepackage[linesnumbered,ruled,vlined]{algorithm2e}
\SetKwComment{Comment}{$\rhd$}{}
\AtBeginDocument{%
  \providecommand\BibTeX{{%
    \normalfont B\kern-0.5em{\scshape i\kern-0.25em b}\kern-0.8em\TeX}}}

\newcommand{\ie}{\textit{i.e.}}
\newcommand{\eg}{\textit{e.g.}}
\newcommand{\red}{\textcolor{red}}
\newcommand{\blue}{\textcolor{blue}}

\newcommand{\purple}{\textcolor{purple}}
\newcommand{\myon}{\purple{\faToggleOn}}
\newcommand{\myoff}{\purple{\faToggleOff}}
\newcommand{\myXSolidBrush}{\blue{\XSolidBrush}}
\newcommand{\myCheckmark}{\purple{\Checkmark}}

\definecolor{veronica-red}{RGB}{196,30,58}

\title{A Survey of Quantized Graph Representation Learning: \\Connecting Graph Structures with Large Language Models}

\author{Qika Lin$^1$, Zhen Peng$^2$, Kaize Shi$^3$, Kai He$^1$, Yiming Xu$^2$, Jian Zhang$^2$,\\
{
\bf Erik Cambria$^4$, Mengling Feng$^1$ }\\
$^1$National University of Singapore\;\;$^2$Xi'an Jiaotong University\\
$^3$University of Southern Queensland\;\;$^4$Nanyang Technological University\\
  \texttt{qikalin@foxmail.com}\quad\texttt{ephfm@nus.edu.sg}
  }

\begin{document}
\maketitle
\begin{abstract}
Recent years have witnessed rapid advances in graph representation learning, with the continuous embedding approach emerging as the dominant paradigm.
    However, such methods encounter issues regarding parameter efficiency, interpretability, and robustness.
    % However, this type of method would face the problems of parameter efficiency, explainability, and robustness.
    %Owing to this, in the recent few years, 
    As a promising solution, Quantized Graph Representation (QGR) learning has recently gained increasing interest, which encodes graph structures with discrete codes instead of conventional continuous embeddings.
    % Quantized Graph Representation (QGR) learning has received more and more attention because of its merits, which represent the graph structure into discrete codes rather than continuous embedding. 
    % , including embedding parameter efficiency, explainability and interoperability, robustness and generalization, and seamless integration with NLP models.
    Given its analogous representation form to natural language, QGR also possesses the capability to seamlessly integrate graph structures with large language models (LLMs).
    % Besides, QGR holds the potentials for seamlessly integrating graph structures with advanced large language models (LLMs), as it has the similar representation form with natural language.
    As this emerging paradigm is still in its infancy yet holds significant promise, we undertake this thorough survey to promote its rapid future prosperity.
    % , aiming to give a current comprehensive picture of it and inspire future research.
    % Considering this type of technique is very promising and in a primary stage, we make a comprehensive review of QGR learning, from the perspectives of general framework, quantized methods, distinctive designs, and applications, hoping give a current comprehensive picture of this new paradigm.
    We first present the background of the general quantization methods and their merits.
    % Moreover, we provide an in-depth demonstration of current QGR studies from the perspectives of quantized strategies, training objectives, distinctive designs, knowledge graph quantization, and applications.
    An in-depth demonstration of existing QGR studies is then carried out from the perspectives of quantized strategies, training objectives, distinctive designs, knowledge graph quantization, and applications.
    Furthermore, we explore the strategies for code dependence learning and integration with LLMs.
    % At last, we give discussions and conclude future directions, aiming to provide a comprehensive picture of QGR and inspire future research.
    The final discussion on strengths and limitations aims to provide a comprehensive picture of QGR and inspire future research directions.
    % This type of method proposes a promising potential of comprehensively modeling graphs with advanced techniques of LLMs.
    % The relevant studies are constantly updated online at: \blue{github.com/}.
\end{abstract}

\section{Introduction}

\begin{figure}[t]
\centering
\includegraphics[width=0.9\linewidth]{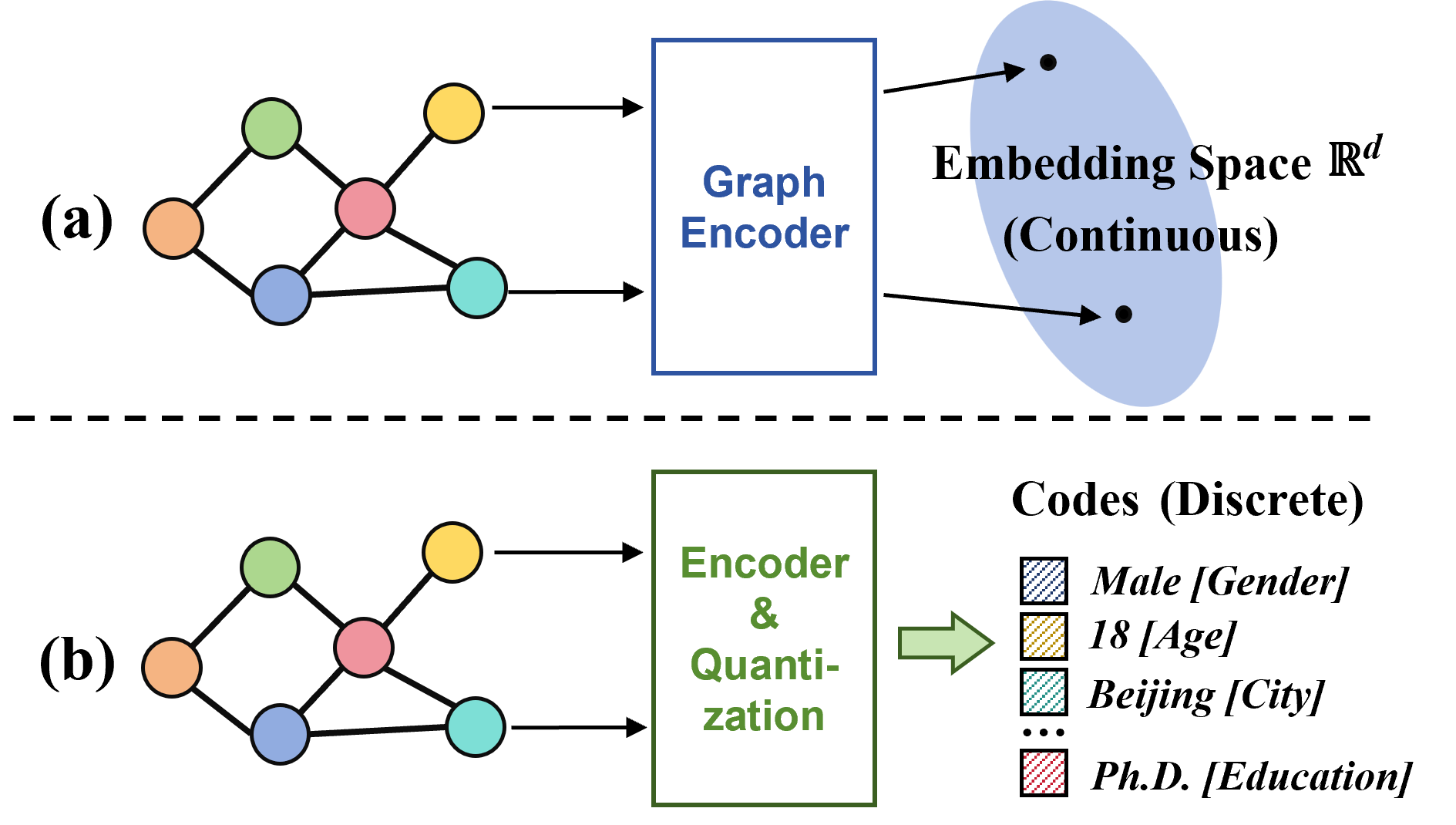}
% \vspace{-0.3cm}
\setlength{\abovecaptionskip}{0.1cm}
\setlength{\belowcaptionskip}{-0.3cm}
\caption{Illustration of different strategies for graph representations. (a) is the continuous graph representation. (b) is the quantized graph representation which represents the graph structure with discrete codes instead of conventional continuous embeddings.}
\label{fig_intro}
\end{figure}

Graph representation usually involves converting nodes, edges, or structures within various graph data into low-dimensional dense embeddings, which is called continuous graph representation~\cite{xia2021graph,lin2021contrastive} as shown in Figure~\ref{fig_intro} (a).
The learned representation is obligated to preserve both node attributes and topological structures~\cite{DBLP:journals/corr/abs-2006-04131}.
% into a form that can be processed by computers, commonly achieved through vector or matrix representation.
% The essential aspect of this process is to preserve the original graph's structural information while compacting it into a lower dimensional space.
% whose key is to maintain the structural information of the original graph in a low-dimensional space.
% In recent years, the field of graph representation learning has experienced significant growth and advancement.
% The field of graph representation learning
This field has made great progress in recent years, and the emergence of numerous advanced technologies has invariably drawn significant attention within the graph community, becoming focal points of extensive studies, \eg, the well-known node2vec~\cite{DBLP:conf/kdd/GroverL16} model, graph neural networks (GNNs)~\cite{DBLP:conf/iclr/KipfW17}, and self-supervised graph learning~\cite{liu2022graph}.
They have achieved significant empirical success in graph-related domains, including social networks, recommendation systems, and anomaly detection~\cite{xia2021graph,DBLP:journals/tkde/LiuJPZZXY23}.
% bioinformatics

\begin{figure*}[t]
\centering
\includegraphics[width=0.9\linewidth]{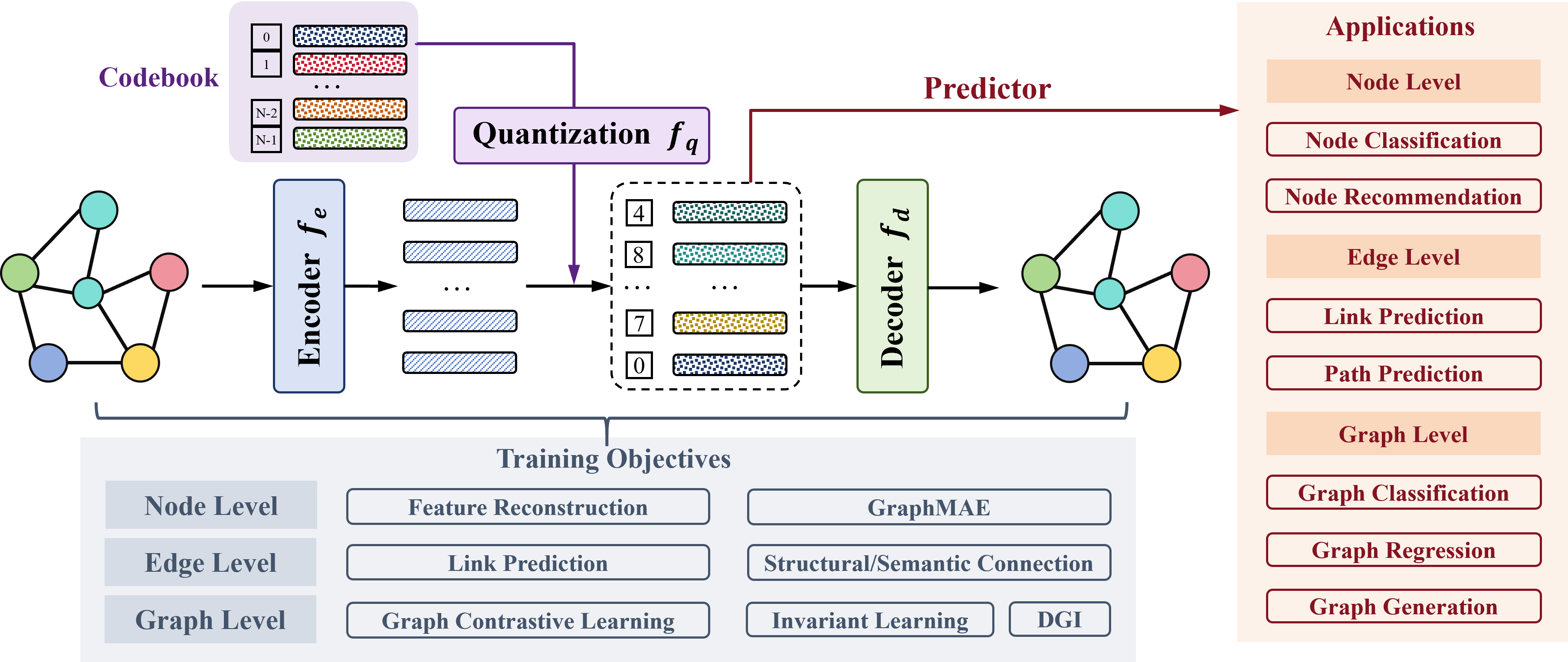}
% \vspace{-0.3cm}
% \setlength{\abovecaptionskip}{-0.1cm}
\setlength{\belowcaptionskip}{-0.3cm}
\caption{The general framework of the QGR studies, which mainly comprises an encoder, decoder, and quantization process.
Training objectives of different levels can be utilized. 
By combining a predictor, multiple applications can be realized.}
\label{fig_arc}
\end{figure*}

% Currently, continuous graph representation is the mainstream direction of graph learning, which is a way to represent graph data in a continuous space, transforming nodes, edges, and graph structures into continuous and real value vector representations.
% Continuous graph representation can retain rich information about the original graph data and identify relationships within it, applicable to a wide range of machine learning and deep learning tasks.
Although continuous graph representation successfully extracts rich semantic information from the input graph to adapt to a wide range of machine learning and deep learning tasks, concerns may arise regarding the embedding efficiency, interpretability, and robustness~\cite{wang2024learning}.
% However, there would be some issues about embedding efficiency, interpretability, and robustness.
Specifically, with the great growth of graph scales, there will be a linear increase in embedding parameters, becoming a considerable challenge in graph learning contexts involving millions or even more nodes.
Besides, continuous embeddings are generated in a black-box fashion and the meaning of the overall representation or any one dimension is unknowable, leading to a lack of interpretability.
Moreover, these continuous representations are typically task-specific and show no significant abstraction at the representation level, making them not robust enough to generalize to a variety of practical tasks as effectively as large language models (LLMs)~\cite{DBLP:journals/tist/NaveedKQSAUABM25}.
% making them insufficiently robust to be effectively generalized to other tasks.

% For example, as shown in Figure, a person represented by embedding is general and ambiguous for human understanding.
% The features of a person, like age (\eg, 18) and gender (male or female) are usually discrete, 
% Furthermore, when facing graph-text integration scenarios, fusing continuous embedding directly into NLP model would be at the cost of losing semantic representation.

% In reality, it is usually preferred to use discrete features for representation.
% instead of continuous embeddings.
Interestingly, in real life, discrete features are often preferred for representation.
For example, as shown in Figure~\ref{fig_intro} (b), descriptions affiliated to an individual are often a combination of discrete and meaningful features, such as age (\eg, \emph{18}), gender (\emph{male} or \emph{female}), and place of residence (\eg, \emph{Beijing}).
This scheme is simple and easy to understand.
% a person represented by embedding is general and ambiguous for human understanding.
% The features of a person, like age (\eg, 18) and gender (male or female) are usually discrete, 
% Furthermore, when facing graph-text integration scenarios, fusing continuous embedding directly into NLP model would be at the cost of losing semantic representation.
Based on this intuition, quantized representation came into being, which is now a hot topic in the computer vision domain that can facilitate image generation by combining with Transfomer or LLMs~\cite{esser2021taming,DBLP:conf/iclr/0004KCY24}.
% The famous XXX model all uses this technique.

Inspired by general quantized representation techniques, quantized graph representation (QGR) learning has recently been proposed to address the aforementioned shortcomings of continuous graph learning.
The core idea is to learn discrete codes for nodes, subgraphs, or the overall graph structures, instead of employing continuous embeddings.
As shown in Figure~\ref{fig_intro} (b), a node can be represented as discrete codes of (\emph{male}, \emph{18}, \emph{Beijing}, \emph{Ph.D.}).
In QGR methods, codes can be learned by differentiable optimization with specific graph targets or self-supervised strategies, illustrated in Figure~\ref{fig_arc}.
It has the capacity to learn more high-level representations due to a compact latent space, which would enhance the representation efficiency, interpretability, and robustness.
% Moreover, when facing graph-text scenarios, the incorporation of continuous embeddings into advanced LLMs may present difficulties, potentially leading to a loss of semantic information.
Recall that when faced with graph-text scenarios, incorporating continuous embeddings into advanced LLMs is inherently challenging and prone to introducing the loss of semantic information. In contrast, due to the consistency with the discrete nature of natural language, the output of QGR would be seamlessly integrated into LLMs, like early-fusion strategy~\cite{DBLP:journals/corr/abs-2405-09818}.
% which holds great potential .
Considering that this emerging paradigm is still in its initial stage but carries substantial potential in the era of LLMs, we undertake this thorough survey.
It encompasses various viewpoints, including different graph types, quantized strategies, training objectives, distinctive designs, applications and so forth.
% We also provide discussions and conclude future directions,
We also conclude future directions,
intending to motivate future investigation and contribute to the advancement of the graph and natural language community.
% The research is in its early stages.
% In this paper, we give a comprehensive survey of related studies from multiple perspectives, expecting to bring awareness of QGR to the community and promote its further development.

To start with, we first describe the general quantization method
% including the merits of QGR and general quantization methods 
in \S\ref{sec_quantization}. Then, we provide a comprehensive introduction to QGR frameworks from multiple perspectives
% of QGR framework, training objectives, distinctive designs, knowledge graph quantization, and application scenarios 
in \S\ref{sec_quan}.
Next, we present the code dependence learning as well as the integration with LLMs in \S\ref{sec_code_dep} and \S\ref{sec_llm}.
% , respectively.
% Finally, we give the discussions and future directions in Section 6.
Finally, we give future directions in \S\ref{sec_future}.

\section{General Quantization Methods}
\label{sec_quantization}

We provide a detailed summary of the merits of QGR in Appendix \S\ref{sec_merits}.
The mainstream quantization methods can be roughly categorized into product quantization, vector quantization, finite scalar quantization, and anchor selection and assignment.

\noindent\textbf{Product quantization (PQ).}
The core idea of PQ~\cite{jegou2010product} is to divide a high-dimensional space into multiple low-dimensional subspaces and then quantize them independently within each subspace. Specifically, it splits a high-dimensional vector into multiple smaller subvectors and quantizes each subvector separately. In the quantization process, each subvector is mapped to a set of a finite number of center points by minimizing the quantization error.

% \red{The VQ-VAE model~\cite{DBLP:conf/nips/OordVK17} is originally proposed for modeling continuous data distribution, such as images, audio and video. It encodes observations into a sequence of discrete latent variables, and reconstructs the observations from these discrete variables. Both encoder and decoder use a shared codebook.
% }
% Taming transformers~\cite{esser2021taming} VQGAN:

\noindent\textbf{Vector Quantization (VQ).}
To effectively learn quantized representations in a differentiable manner,
the VQ strategy~\cite{DBLP:conf/nips/OordVK17,esser2021taming} is proposed and has emerged as the predominant method in the field.
The core idea involves assigning the learned continuous representations to the codebook's index and employing the Straight-Through Estimator~\cite{DBLP:journals/corr/BengioLC13} for effective optimization.
The formulation is as follows.
$f_e$, $f_d$, $f_q$ denote the encoder, decoder, and quantization process, respectively.
$\mathbf{C}\in \mathbb{R}^{M\times d}$ is the codebook representation with dimension $d$, which corresponds to $M$ discrete codewords $\mathcal{C}=\{1, 2, \cdots, M\}$.
$\mathcal{G}=\{\mathcal{V}, \mathcal{E}, \mathbf{X}, \mathbf{Y}\}$ is the graph, where $\mathcal{V}$ is the node set with size $n$ and $\mathcal{E}$ is the edge set.
The presence of $\mathbf{X}$ or $\mathbf{Y}$ is not guaranteed, where $\mathbf{X}\in \mathbb{R}^{n\times d}$ is the input feature of a graph and $\mathbf{Y} = \{y_1, y_2, \cdots\}$ is the label set of each node.
% $\mathbf{X}\in \mathbb{R}^{n\times d}$ is the input feature of a sample graph with $n$ nodes, and the feature dimension is $d$.
% $\mathbf{X}\in \mathbb{R}^{n\times d}$ is the input feature of a simple.
$\widetilde{\mathbf{X}}\in \mathbb{R}^{n\times d}$ is the latent representation after encoder and $\widehat{\mathbf{X}}\in \mathbb{R}^{n\times d}$ is the reconstruction feature after quantization and decoder, \ie,
$\widetilde{\mathbf{X}}=f_e(\mathbf{X})$,
$\widehat{\mathbf{X}}=f_d(f_q(\widetilde{\mathbf{X}}))$.

Specifically, the VQ strategy yields a nearest-neighbor lookup between latent representation and prototype vectors in the codebook for quantized representations:
\begin{equation}
\small
\label{eq_vq}
f_q(\widetilde{\mathbf{X}}_i)=\mathbf{C}_{m},\;m=\underset{j\in \mathcal{C}}{\mathop{\arg\min}}\| \widetilde{\mathbf{X}}_i-\mathbf{C}_{j}\|_2^2,
\end{equation}
indicating that node $i$ is assigned codeword $c_m$.
Further, the model can be optimized by the Straight-Through Estimator:
\begin{equation}
\small
  \mathcal{L}_{vq}=\frac{1}{n}\sum_{i=1}^{n}\underbrace{\|\text{sg}[\widetilde{\mathbf{X}}_i]-\mathbf{C}_{m}\|}_{codebook}+
  \underbrace{\beta\|\text{sg}[\mathbf{C}_{m}]-\widetilde{\mathbf{X}}_i\|}_{commitment},
\end{equation}
where \text{sg} represents the stop-gradient operator. $\beta$ is a hyperparameter to balance and is usually set to 0.25.
In this loss, the first codebook loss encourages the codeword vectors to align closely with the encoder's output.
The second term of the commitment loss aids in stabilizing the training process by encouraging the encoder's output to adhere closely to the codebook vectors, thereby preventing excessive deviation.

% \paragraph{Residual Vector Quantization (RVQ).}
% \red{Residual quantization is actually a multi-stage quantization method, and the algorithm flow is very simple. As shown in the figure below, the codebook stores the residual values of the quantized vector and the original vector at each step, and the original vector can be approximated infinitely by quantizing the residual values at multiple steps.
% Is to choose a CodeBook vector to fit Residual, but this fitting will still have errors, then fit this "fitting error", gradually recursive down.}

\begin{table*}[t!]
\setlength\tabcolsep{3pt} 
% \small
\setlength{\abovecaptionskip}{0.03cm}
\setlength{\belowcaptionskip}{-0.5cm}
\caption{The summary of QGR learning studies.
Abbreviations are as follows: HOG: homogeneous graph that includes attributed graph (AG), text-attributed graph (TAG), and molecular graph (MG), HEG: heterogeneous graph.
% multiple relations?
For training and applications, there are STC: structure connection, SEC: semantic connection, NC: node classification, NR: node recommendation, FR: feature reconstruction, LP: Link Prediction, PP: path prediction, GC: graph classification, GR: graph regression, GG: graph generation, MCM: masked code modeling, GCL: graph contrastive learning, NTP: next-token prediction, IC: invariant learning, QA: question answering.
\myoff\;and \myon\; indicate for one-stage or multi-stage learning, respectively.
% code learning in one step or multi-steps.
% method includes the information for supervised. or unsupervised.
In the training column, ``/''split the training stages.
% Self-Sup means the self-supervised training target and is task-agnostic.
    }
    \centering
    \resizebox{1.0\textwidth}{!}{
    \begin{tabular}{l|ccllllcc}
        \toprule
        Model  &one/p&Self-Sup& Graph Type &  Method & Training & Application&Dependence&w/ LM\\
        \midrule
        SNEQ~\shortcite{he2020sneq}&\myoff&\myCheckmark&AG (HOG)&PQ&STC, SEC& NC, LP, NR&\myXSolidBrush&\myXSolidBrush \\
        d-SNEQ~\shortcite{DBLP:journals/tnn/HeGSL23}&\myoff&\myCheckmark&AG (HOG)&PQ&STC, SEC& NC, LP, NR, PP&\myXSolidBrush&\myXSolidBrush \\
        Mole-BERT~\shortcite{DBLP:conf/iclr/XiaZHG0LLL23}&\myon&\myCheckmark&MG (HOG)&VQ&FR/MCM, GCL&GC, GR&\myCheckmark&\myXSolidBrush\\
        iMoLD~\shortcite{DBLP:conf/nips/ZhuangZDBWLCC23}&\myoff&\myCheckmark&MG (HOG)&VQ&IL, GC (GR)&GC, GR&\myXSolidBrush&\myXSolidBrush\\
        VQGraph~\shortcite{yang2024vqgraph} &\myon&\myCheckmark& HOG & VQ&FR, LP/NC, Distill&  NC&\myXSolidBrush&\myXSolidBrush\\
        Dr.E~\shortcite{liu2024dr}&\myon& \myCheckmark & TAG (HOG)&RVQ&FR, LP, NC/FR, NTP&  NC&\myCheckmark&\myCheckmark  \\
        STAG~\shortcite{DBLP:conf/kdd/Bo0025}&\myon& \myCheckmark & TAG (HOG)&VQ& GraphMAE, Contrast/&  NC&\myXSolidBrush&\myCheckmark  \\
        DGAE~\shortcite{DBLP:journals/tmlr/BogetGK24}&\myon&\myCheckmark& HEG &VQ&LP/NTP&  GG&\myCheckmark&\myXSolidBrush\\
        LLPS~\shortcite{DBLP:journals/corr/abs-2405-15840}&\myoff&\myCheckmark& MG (HOG)&FSQ&Reconstruction&  GG&\myXSolidBrush&\myXSolidBrush\\
        GLAD~\shortcite{boget2024glad}         &\myon&\myCheckmark& MG (HOG) &FSQ&FR+LP/Diffusion Bridge&  GG&\myCheckmark&\myXSolidBrush\\
        Bio2Token~\shortcite{liu2024bio2token}&\myoff&\myCheckmark&MG (HOG)&FSQ&FR&GG&\myXSolidBrush&\myXSolidBrush\\
        GQT~\shortcite{wang2024learning}&\myon&\myCheckmark&HOG&RVQ&DGI, GraphMAE2/NC&NC&\myXSolidBrush&\myXSolidBrush\\
        NID~\shortcite{DBLP:journals/corr/abs-2405-16435}&\myon&\myCheckmark&HOG&RVQ&GraphMAE (GraphCL)/&NC, LP, GC, GR, etc&\myXSolidBrush&\myXSolidBrush\\
        UniMoT~\shortcite{DBLP:journals/corr/abs-2408-00863}&\myon&\myCheckmark&MG (HOG)&VQ&FR/NTP&GC, GR, Captioning, etc.&\myXSolidBrush&\myCheckmark\\
        % MSPmol~\shortcite{DBLP:conf/ijcnn/LuPZC24}&\myon&\myCheckmark&MOG&VQ&\red{NC, edge type, etc}&GC&\myXSolidBrush&\myXSolidBrush\\
         NodePiece~\shortcite{DBLP:conf/iclr/0001DWH22}&\myon&\myXSolidBrush&KG&ASA&--/LP&LP&\myXSolidBrush&\myXSolidBrush\\
         EARL~\shortcite{DBLP:conf/aaai/ChenZYZGPC23}&\myon&\myXSolidBrush&KG&ASA&--/LP&LP&\myXSolidBrush&\myXSolidBrush\\
         RandomEQ~\shortcite{DBLP:conf/emnlp/LiWLZM23}&\myon&\myXSolidBrush&KG&ASA&--/LP&LP&\myXSolidBrush&\myXSolidBrush\\
         SSQR~\shortcite{lin2024self}&\myon&\myCheckmark&KG&VQ&LP, Distilling/NTP&LP, Triple Classification&\myXSolidBrush&\myCheckmark\\
         ReaLM~\shortcite{DBLP:journals/corr/abs-2510-09711}&\myon&\myCheckmark&KG&RVQ&FR/NTP&LP, Triple Classification&\myXSolidBrush&\myCheckmark\\
         MedTok~\shortcite{DBLP:conf/icml/0001MHJFGSZ25}&\myon&\myCheckmark&KG&VQ&Contrast/&Classification, QA&\myXSolidBrush&\myCheckmark\\
        \bottomrule
    \end{tabular}
    }
    \label{tab_all_studies}
\end{table*}

Using VQ techniques, a continuous vector can be quantized to the counterpart of a discrete code.
However, there would be distinctions and differences in the vector pre and post-quantization.
To address it, Residual Vector Quantization (RVQ)~\cite{DBLP:conf/cvpr/LeeKKCH22} is proposed to constantly fit residuals caused by the quantization process.
RVQ is actually a multi-stage quantization method and usually has multiple codebooks.
The codebook stores the residual values of the quantized vector and the original vector at each step, and the original vector can be approximated infinitely by quantizing the residual values at multiple steps.
The whole process is shown in Algorithm~\ref{algorithm_rvq} of the Appendix.

\noindent\textbf{Finite Scalar Quantization (FSQ).}  
Because of the introduction of the codebook,
there would be complex technique designs and codebook collapse in VQ techniques.
Thus, FSQ~\cite{DBLP:conf/iclr/MentzerMAT24} introduces a very simple quantized strategy, \ie, directly \emph{rounding} to integers rather than explicitly introducing parameters of the codebook.
FSQ projects the hidden representations into a few dimensions (usually fewer than 10) and each dimension is quantized into a limited selection of predetermined values, which inherently formulates a codebook as a result of these set combinations.
It is formulated as:
\begin{equation}
\small
\text{FSQ}(\widetilde{\mathbf{X}}_i)=\mathcal{R}[(L-1)\sigma(\widetilde{\mathbf{X}}_i)]\in \{0,1,\cdots,L-1\}^d,
\end{equation}
where $\mathcal{R}$ is the rounding operation and $L\in \mathbb{N}$ is the number of unique values.
The process of rounding simply adjusts a scalar to its nearest integer, thereby executing quantization in every dimension.
$\sigma$ is a sigmoid or tanh function.
Its loss is calculated by also using the Straight-Through Estimator:
\begin{equation}
\small
\begin{aligned}
\mathcal{L}_{fsq}=&\frac{1}{n}\sum_{i=1}^{n}(L-1)\sigma(\widetilde{\mathbf{X}}_i)\\
&+\text{sg}[\mathcal{R}[(L-1)\sigma(\widetilde{\mathbf{X}}_i)]-(L-1)\sigma(\widetilde{\mathbf{X}}_i)].
\end{aligned}
\end{equation}
FSQ has the advantages of better performance, faster convergence and more stable training,
particularly when dealing with large codebook sizes.

\noindent\textbf{Anchor Selection and Assignment (ASA).} It is typically unsupervised and often employs prior strategies to identify informative nodes within the graph, \eg, the Personalized PageRank~\cite{page1999pagerank} and node degree.
Consequently, each node can be attributed to a combination of anchors that exhibit structural or semantic relationships,
like using the shortest path.
This form of quantization designates the code as a real node, unlike the aforementioned three methods.

\section{Quantized Graph Learning Framework}
\label{sec_quan}

We summarize the main studies of QGR in Table~\ref{tab_all_studies} from the primary perspectives of graph types, quantization methods, training strategies, and applications.
Additionally, we also highlight if there are in pipeline and self-supervised manners, as well as whether they learn code dependence and integrate with language models.

The general process of QGR is illustrated in Figure~\ref{fig_arc}, which comprises the encoder $f_e$, decoder $f_d$, and quantization process $f_q$.
The encoder $f_e$ is to model the original graph structure into latent space, where MLPs~\cite{he2020sneq} and GNNs~\cite{DBLP:conf/iclr/XiaZHG0LLL23} are usually utilized.
The decoder $f_d$ also commonly utilizes GNNs for structure embedding and the quantization process $f_q$ implements the strategies outlined in Section~\ref{sec_quantization}.
In particular, Bio2Token~\cite{liu2024bio2token} utilizes Mamba~\cite{DBLP:journals/corr/abs-2312-00752} as both encoder and decoder.
UniMoT~\cite{DBLP:journals/corr/abs-2408-00863} leverages the pre-trained MoleculeSTM molecule encoder~\cite{DBLP:journals/natmi/LiuNWLQLTXA23} to connect the molecule graph encoder and causal Q-Former~\cite{DBLP:conf/icml/0008LSH23}.
In the following sections, we will delve into the details of the model design.

% encoder and decoder

% encoder, MLP (SNEQ)  GNN(Mole-BERT)

% \red{Homophilous graphs are characterized by nodes with similar classes being connected to each other, whereas heterophilous graphs exhibit connections between nodes with different classes.}

% Proprocessing:
% UniMoT
% \red{We connect the molecule encoder and Causal Q-Former, leveraging the pretrained MoleculeSTM molecule encoder~\cite{DBLP:journals/natmi/LiuNWLQLTXA23}. The molecule encoder remains frozen while only the Causal Q-Former is updated.
% }
% \red{We connect the Causal Q-Former with subsequent blocks and use the objective defined in Equation (2). We employ the pretrained ChemFormer~\cite{DBLP:journals/mlst/IrwinDHB22} as the generative model. Specifically, we leverage the SMILES encoder and the SMILES decoder provided by ChemFormer.}

\subsection{Quantized Strategies}

Except for SNEQ~\cite{he2020sneq} and d-SNEQ~\cite{DBLP:journals/tnn/HeGSL23}, the majority of research tends to employ VQ-related strategies.
Beyond the general techniques in VQ,
Dr.E~\cite{liu2024dr} utilizes RVQ in one specific layer, which is called intra-layer residue.
For more effective in encoding the structural information of the central node, it involves preserving the multi-view of a graph.
Specifically, between GCN layers, the inter-layer residue is carried out to enhance the representations:
\begin{equation}
\small
\mathbf{h}_v^{l+1}=\sigma\big(\mathbf{W}\cdot\mathrm{Con}(\mathbf{h}_v^l+\mathrm{Pool}(\{\mathbf{c}_{v,k}^l\}_{k=1}^K),\mathbf{h}_{\mathcal{N}(v)}^l)\big),
\end{equation}
where $\mathbf{h}$ and $\mathbf{c}$ denote the latent representations and quantized ones.
$K$ is the number of learned codes for each layer.
Using this process, the discrete codes of each node are generated in a gradual (auto-regressive) manner, following two specific orders: from previous to subsequent, and from the bottom layer to the top.
This approach guarantees comprehensive multi-scale graph structure modeling and efficient code learning.
iMoLD~\cite{DBLP:conf/nips/ZhuangZDBWLCC23} proposes another type of \emph{Residual} VQ, incorporating both the continuous and discrete representations, \ie, $\widetilde{\mathbf{X}}_i+f_q(\widetilde{\mathbf{X}}_i)$, as the final node representation.

GLAD~\cite{boget2024glad} and Bio2Token~\cite{liu2024bio2token} utilize the straightforward approach, FSQ, to facilitate the learning of discrete codes.
For KG quantization, early methods such as NodePiece~\cite{DBLP:conf/iclr/0001DWH22}, EARL~\cite{DBLP:conf/aaai/ChenZYZGPC23}, and RandomEQ~\cite{DBLP:conf/emnlp/LiWLZM23} have adopted the unsupervised learning approach ASA to achieve effective learning outcomes.
For more condensed code representations, SSQR~\cite{lin2024self} employs the VQ strategy to capture both the structures and semantics of KGs.

\subsection{Training Objectives}

In this section, we will delve deeper into training details from the perspectives of node, edge, and graph levels.

\noindent\textbf{Node Level.}
Feature reconstruction usually serves as a simple self-supervised method for QGR learning, which is to compare the original node features with reconstructed features based on the learned codes.
It mainly has the following two implementation types:
\begin{equation}
\small
    \mathcal{L}_{fr}=\frac{1}{n}\sum_{i=1}^n\big(1-\frac{\mathbf{X}_i^{\top}\widehat{\mathbf{X}}_i}{\|\mathbf{X}_i\|\|\widehat{\mathbf{X}}_i\|}\big)^{\gamma},\mathcal{L}_{fr}=\frac{1}{n}\sum_{i=1}^n\|\mathbf{X}_i-\widehat{\mathbf{X}}_i\|^2,
\end{equation}
The first involves employing cosine similarity with scaled parameter $\gamma\geq 1$, as implemented in Mole-BERT and VQGraph.
The second is the mean square error as in Dr.E.
NID~\cite{DBLP:journals/corr/abs-2405-16435} utilizes GraphMAE~\cite{DBLP:conf/kdd/HouLCDYW022} for self-supervised learning, which can be viewed as a masked feature reconstruction.
It involves selecting a subset of nodes, masking the node features, encoding by a message-passing network, and subsequently reconstructing the masked features with a decoder.
 
\noindent\textbf{Edge Level.}
It is a primary strategy to acquire the graph structures in QGR codes using a self-supervised paradigm.
The direct link prediction, \ie, edge reconstruction, is a widely adopted technique to assess the presence or absence of each edge before and after the reconstruction process as:
\begin{equation}
\small
\begin{aligned}
    &\mathcal{L}_{lp}=\|\textbf{A}-\sigma(\widehat{\mathbf{X}}\cdot \widehat{\mathbf{X}}^{\top}) \|^2,\\
    \mathcal{L}_{lp}=-\frac{1}{|\mathcal{E}|}&\sum_{j=1}^{|\mathcal{E}|}\big[y_j\log(\hat{y}_j)+(1-y_j)\log(1-\hat{y}_j)\big],\\
\end{aligned}
\end{equation}
where the mean square error (\eg, VQGraph) and the binary cross-entropy are utilized to optimize (\eg, Dr.E), respectively.
$\hat{y}_j$ is the prediction of probability for the existence of the edge based on the learned representations.
$y_j\in\{0,1\}$ is the ground label.

Beyond direct edge prediction, \citet{he2020sneq} proposes high-level connection prediction between two nodes, \ie, structural connection, and semantic connection.
% Structural connection between two nodes, which 
The former aims to accurately maintain the shortest distances $\delta$  within the representations:
\begin{equation}
\small
    \mathcal{L}_{stc}=\frac{1}{n}\sum_{(i,j,k)}^n \text{max}\big(D_{i,j}-D_{i,k}+\delta_{i,j}-\delta_{i,k},0\big),
\end{equation}
where $D$ is to calculate the distance of two nodes based on the learned codes.
% in the embedding space.
% $\delta$ indicates the shortest distance of two nodes. 
Considering that nodes sharing identical labels ought to be situated closer together within the embedding space,
semantic connection is proposed:
\begin{equation}
\small
    \mathcal{L}_{sec}=\frac{1}{n\cdot T}\sum_{i}^{n}\sum_{i,j=1}^{T}(D_{i,j}-S_{i,j})^2,
\end{equation}
where $T$ is the sample size. $S$ presents the constant semantic margin, which is set to 0 if the labels of two nodes are totally distinct, otherwise, it is set to a constant.

\noindent\textbf{Graph Level.}
For high-level representations, the graph modeling targets are proposed.
For instance, graph contrastive learning~\cite{DBLP:journals/corr/abs-1807-03748} is a widely utilized method for modeling relationships between graph pairs:
\begin{equation}
\small
\mathcal{L}_{gcl}=-\frac{1}{|\mathcal{D}|}\sum_{\mathcal{G}\in\mathcal{D}}\log\frac{e^{sim(\mathbf{h}_{g1},\mathbf{h}_{g2})/\tau}}{\sum_{\mathcal{G}'\in\mathcal{B}}e^{sim(\mathbf{h}_{g1},\mathbf{h}_{\mathcal{G}'})/\tau}},
\end{equation}
where $\mathcal{D}$ is the dataset that contains all graphs and $\mathbf{h}_{g}$ is the graph representation based on $f_q(\widetilde{\mathbf{X}})$.
$g1$ and $g2$ can be seen as positive samples for the target graph, which are usually transformed by the target graph $\mathcal{G}$.
For example, Mole-BERT~\cite{DBLP:conf/iclr/XiaZHG0LLL23} utilizes different masking ratios (\eg,  15\% and 30\%) for molecular graphs to obtain $g1$ and $g2$.
$\mathcal{B}$ is contrastive samples,  typically derived from the sampled batch.
GraphCL~\cite{DBLP:conf/nips/YouCSCWS20} is also a well-known framework of this kind based on graph augmentations, including node dropping, edge perturbation, attribute masking, and subgraph, which is used by NID~\cite{DBLP:journals/corr/abs-2405-16435}.

Furthermore,
motivated by a straightforward self-supervised learning framework that identifies varied augmentation views as similar positive pairs, iMoLD~\cite{DBLP:conf/nips/ZhuangZDBWLCC23} considers the learned invariant and global aspects (invariant plus spurious) as positive pairs, aiming to enhance their similarity.
An MLP-based predictor is introduced to transform the output of one view and align it with the other.
The whole process can be interpreted as \emph{invariant learning}.
Deep Graph Infomax (DGI)~\cite{DBLP:conf/iclr/VelickovicFHLBH19} can also be used as a high-level QGR learning method, which is a contrastive approach that compares local (node) representations with global (graph or sub-graph) ones.
% whereas GMAE2 combines generative and distillation objectives to jointly reconstruct masked features and track teacher representations.
GQT~\cite{wang2024learning} utilize both high-level DGI~\cite{DBLP:conf/iclr/VelickovicFHLBH19} and node-level GraphMAE2~\cite{DBLP:conf/www/HouHCLDK023} as training targets.
% GMAE2 combines generative and distillation objectives to jointly reconstruct masked features and track teacher representations.
% \red{GraphCL: Such approaches usually consider each sample as its own class, that is, a positive pair consists of two different views of it; and all other samples in a batch are used as the negative pairs during training.  Specifically, a minibatch of N graphs is randomly sampled and subjected to contrastive learning. This process results in 2N augmented graphs, along with a corresponding contrastive loss to be optimized.}

\subsection{Distinctive Designs}

\noindent\textbf{Codebook Design.}
Beyond randomly setting the codebook for updating, there are several specifically designed strategies for domain knowledge adaption and generalization.
For example, Dr.E~\cite{liu2024dr} sets the vocabulary of LLaMA as the codebook, realizing token-level alignment between GNNs and LLM.
In this way, each node of graphs can be quantized to a permutation of language tokens and then can be directly input to LLMs to make predictions.
When modeling molecular graphs, Mole-BERT~\cite{DBLP:conf/iclr/XiaZHG0LLL23} categorizes the codebook embeddings into various groups, each representing a distinct type of atom. For instance, the quantized codes of carbon, nitrogen and oxygen are confined to different groups.

\noindent\textbf{Training Pipelines.}
Aside from the single-stage learning process applied to QGR and particular tasks, some methods incorporate multiple stages into the training framework.
GQT~\cite{wang2024learning} modulate the learned codes through hierarchical encoding and structural gating, which are subsequently fed into the Transformer network and aggregate the learned representations through an attention module.
Based on the learned quantized representation, NID~\cite{DBLP:journals/corr/abs-2405-16435} trains an MLP network for downstream tasks, such as node classification and link prediction.

\subsection{KG Quantization}

Recently, several KG quantization methods have been proposed for effective embedding, responding to the increasing demand for larger KGs.
Unlike the techniques employed in the HOG or HEG setting, these methods tend to leverage an unsupervised paradigm for quantization.
NodePiece~\cite{DBLP:conf/iclr/0001DWH22}, EARL~\cite{DBLP:conf/aaai/ChenZYZGPC23}, and random entity quantization (RandomEQ for short)~\cite{DBLP:conf/emnlp/LiWLZM23} are all in such a framework.
They first select a number of entities as anchors and then utilize the structural statistical strategy to match them for each entity.
Specifically, NodePiece employs metrics such as Personalized PageRank~\cite{page1999pagerank} and node degree to select certain anchor entities and subsequently view top-$k$ closest anchors as entity codes.
EARL and RandomEQ sample 10\% entities as anchors and then assign soft weights to each anchor according to the similarity between connected relation sets.
Given the limitations of these methods in thoroughly capturing the structure and semantics of KGs, SSQR~\cite{lin2024self} presents a self-supervised approach that employs learned discrete codes to reconstruct structures and imply semantic text, offering more holistic modeling with only 16 quantized codes.
Furthermore, ReaLM~\cite{DBLP:journals/corr/abs-2510-09711} introduces an RVQ framework that compresses high-dimensional and continuous KG embeddings (pretrained RotatE~\cite{DBLP:conf/iclr/SunDNT19} embeddings) into compact and discrete code sequences while effectively preserving rich relational and semantic information.
More detailed comparison between SSQR and ReaLM are shown in Appendix \S\ref{sec_ssqr_realm}.

\subsection{Application Scenarios}

% \paragraph{Node Level}
% \paragraph{Edge Level.}
% link prediction (LP)
% \paragraph{Graph Level.}
% molecular property prediction in Mole-BERT, as graph classification task.

\noindent\textbf{General Graph Tasks.}
Based on the learned quantized codes, many general graph tasks can be addressed, for example, node classification, link prediction, graph classification, graph regression, and graph generation.
In addition, SNEQ~\cite{he2020sneq} can be employed for node recommendation that ranks all nodes based on a specific distance metric and suggests the nearest node.
d-SNEQ~\cite{DBLP:journals/tnn/HeGSL23} is further utilized for path prediction that predicts the path (\eg, the shortest path) between two nodes.

Based on the learned structure-aware codes for nodes, VQGraph~\cite{yang2024vqgraph} enhances the GNN-to-MLP distillation by proposing a new distillation target, namely \emph{soft code assignments}, which utilizes the Kullback–Leibler divergence to make two distributions (codes distributions by GNN and MLP) be close together.
This can directly transfer the structural knowledge of each node from GNN to MLP.
The results show it can improve the expressiveness of existing graph representation space and facilitate structure-aware GNN-to-MLP distillation.
In the KG scenarios, NodePiece~\cite{DBLP:conf/iclr/0001DWH22}, EARL~\cite{DBLP:conf/aaai/ChenZYZGPC23}, RandomEQ~\cite{DBLP:conf/emnlp/LiWLZM23}, SSQR~\cite{lin2024self} all can be used for KG link prediction, which is a ranking task that to predict the object entity based the given subject entity and the relation, \ie, $(s,r,?)$.
Using the learned codes as features, SSQR can also fix the triple classification problem, predicting the validity of the given triple $(s,r,t)$.

\noindent\textbf{Molecular Tasks \& AI for Science.}
Recently, QGR methods have been widely used for molecular tasks and AI for science scenarios~\cite{chen2024improving}, including molecular property prediction (classification and regression), molecule-text prediction, and protein \& RNA reconstruction.
Specifically, DGAE~\cite{DBLP:journals/tmlr/BogetGK24} conduct the graph generation.
It first iteratively samples discrete codes and then generates the graph structures by the pre-trained decoder, which can be used in the generation for molecular graphs and has the potential for drug design, material design and protein design.
% $P_\theta$
% \red{For generation, we iteratively sample from p$\theta$(ki,c), and get the corresponding zi,c. Sequence generation stops upon sampling the end-of-sequence token or reaching the maximum number of node embeddings nmax. So, we can generate graphs of various sizes, and we need at most nmaxC iterations to generate an instance.}
LLPS~\cite{DBLP:journals/corr/abs-2405-15840} shows the potential for the reconstruction of protein sequences by training a de novo generative model for protein structures using a vanilla decoder-only Transformer model.
% GLAD used for the molecular generation task.
Bio2Token~\cite{liu2024bio2token} conducts efficient representation of large 3D molecular structures with high fidelity for molecules, proteins, and RNA, holding the potential for the design of biomolecules and biomolecular complexes.
UniMoT~\cite{DBLP:journals/corr/abs-2408-00863} unifies the molecule-text applications, including molecular classification \& regression, molecule captioning, molecule-text retrieval, and caption-guided molecule generation.
% It demonstrates the impressive and comprehensive potentials that arise from the combination of QGR and LLMs.

\section{Code Dependence Learning}
\label{sec_code_dep}

Inspired by the distribution learning of the discrete codes in computer vision field~\cite{esser2021taming} that predicts the next code by the auto-regressive Transformer,
QGR methods also implement this technique, facilitating comprehensive semantic modeling and supporting the generation tasks.
Similar to BERT~\cite{DBLP:conf/naacl/DevlinCLT19} pre-training style, MOLE-BERT adopts a strategy similar to Masked Language Modeling (MLM). 
It employs this method to pre-train the GNN encoder by randomly masking certain discrete codes. Subsequently, it pre-trains GNNs to predict these masked codes, a process known as Masked Code Modeling (MCM):
\begin{equation}
\small
\mathcal{L}_{mcm}=-\sum_{\mathcal{G}\in\mathcal{D}}\sum_{j\in\mathcal{M}}\log p(c_j|\mathcal{G}_{\mathcal{M}}),
\end{equation}
where $\mathcal{M}$ is the set of masked nodes in the graph and $c_j$ is the quantized codes.
Auto-regressive code dependence learning is another mainstream strategy (usually using Transformer) based on next token prediction:
\begin{equation}
\small
\mathcal{L}_{ntp}=-\frac{1}{N}\prod_{j=1}^{N} P_\theta(\textbf{\textit{c}}_j|\textbf{\textit{c}}_1,\textbf{\textit{c}}_2,\cdots, \textbf{\textit{c}}_{j-1}),
\end{equation}
where network $P_\theta$ is with parameter $\theta$.
$N$ is the length of the code sequence $\textbf{\textit{c}}$.
DGAE~\cite{DBLP:journals/tmlr/BogetGK24} splits the latent representation of each node into $C$ parts.
So after the quantization, each graph can be represented as $n\times C$ code permutation.
To learn the dependence of it, the 2D Transformer~\cite{DBLP:conf/nips/VaswaniSPUJGKP17} is introduced to auto-regressively generate the codes, \ie, the joint probability can be $\prod_{i=1}^{n}\prod_{j=1}^{C}P_\theta(\textbf{\textit{c}}_{i,j}|\textbf{\textit{c}}_{<i,1},\cdots, \textbf{\textit{c}}_{<i,C},\textbf{\textit{c}}_{i,<j})$.

Moreover, GLAD~\cite{boget2024glad} implements diffusion bridges~\cite{DBLP:conf/iclr/LiuW0l23} for codes' dependence, where Brownian motion is utilized as a non-conditional diffusion process defined by a stochastic differential equation (SDE).
Based on the learned model bridge, discrete practical codes can be acquired by the iterative denoising process from the random Gaussian noise.
Followed by the pre-trained decoder, these codes can be used for the graph generation.
Although Dr.E does not directly or explicitly learn the dependence of the codes, it introduces intra-layer and inter-layer residuals to gradually generate the codes, which enhances the representation of sequential information as the newly generated code depends on the previously obtained codes.
It can be viewed as an implicit auto-regressive manner.

% \subsection{Distribution Learning}

% \subsection{Incorporation with Language Models}

\section{Integration with LLMs}
\label{sec_llm}

\begin{figure}[t]
\centering
\includegraphics[width=0.99\linewidth]{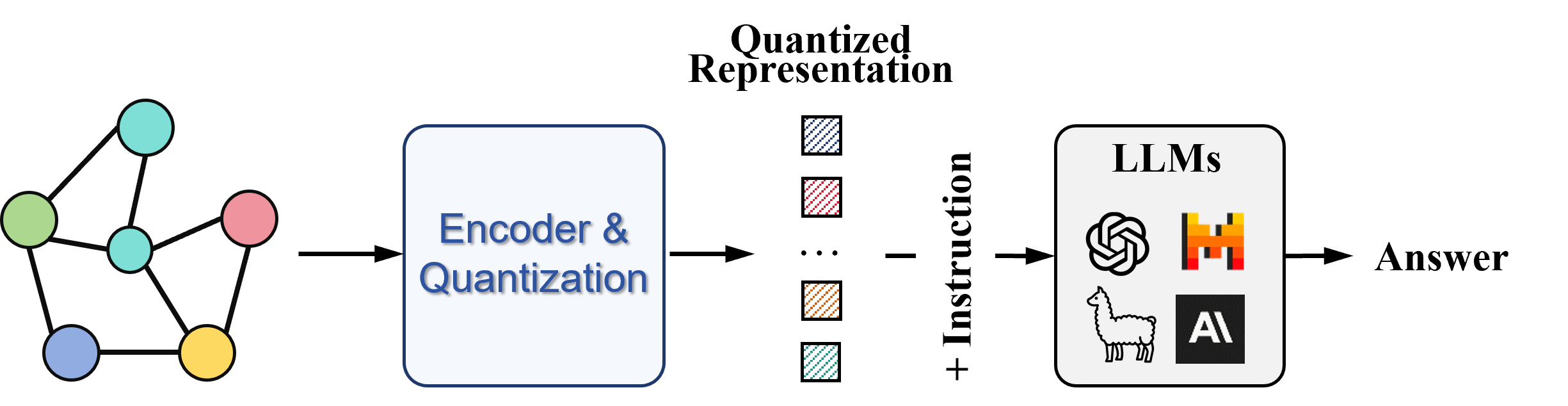}
% \vspace{-0.3cm}
\setlength{\abovecaptionskip}{-0.1cm}
\setlength{\belowcaptionskip}{-0.2cm}
\caption{Illustration of integrating QGR with LLMs.}
\label{fig_llm}
\end{figure}

The quantized codes, being discrete, share a similar structure with natural language.
As such, QGR methods can be seamlessly incorporated with LLMs to facilitate robust modeling and generalization as shown in Figure~\ref{fig_llm}.
Table~\ref{tab_instruction} provides examples of tuning instructions for both Dr.E and SSQR, arranged for better understanding and intuitive interpretation.

\begin{table}[]
\small
    \centering
    \begin{tabular}{c}
                \begin{tcolorbox}[colback=gray!10,%gray background
                      colframe=black,% black frame colour
                      width=7.cm,% Use 5cm total width,
                      boxrule=0.8pt,
                      arc=1mm, auto outer arc,
                      left = 1mm, %文字离线框左边的边距
                        right = 1mm,%同上
                        top = 1mm,%同上
                        bottom = 1mm,%同上
                     ]
                \footnotesize{
                \textbf{Input:} Given a node, you need to classify it among `Case Based', `Genetic Algorithms'.... With the node's 1-hop information being `\blue{amass}', `\blue{traverse}', `\blue{handle}'..., 2-hop information being `\blue{provable}', `\blue{revolution}', `\blue{creative}'..., 3-hop information being `\blue{nous}', `\blue{hypothesis}', `\blue{minus}', the node should be classified as:\\
                \textbf{Output:} Rule Learning
                }
                \end{tcolorbox} \\
                (a) Node classification, taken from Dr.E~\cite{liu2024dr}.\\
                \begin{tcolorbox}[colback=gray!10,%gray background
                      colframe=black,% black frame colour
                      width=7.cm,% Use 5cm total width,
                      boxrule=0.8pt,
                      arc=1mm, auto outer arc,
                      left = 1mm, %文字离线框左边的边距
                        right = 1mm,%同上
                        top = 1mm,%同上
                        bottom = 1mm,%同上
                     ]
                \footnotesize{
                \textbf{Input:}  Given a triple in the knowledge graph, you need to predict its validity based on the triple itself and entities' quantized representations.\\
                The triple is: (\emph{h}, \emph{r}, \emph{t})\\
                The quantized representation of entity \emph{h} is: \blue{[Code(\emph{h})]}\\
                The quantized representation of entity \emph{t} is: \blue{[Code(\emph{t})]}\\
                Please determine the validity of the triple and respond True or False.\\
                \textbf{Output:} True/False
                }
                \end{tcolorbox} \\
                (b) KG triple classification, from SSQR~\cite{lin2024self}.\\
    \end{tabular}
    \setlength{\abovecaptionskip}{0.09cm}
    \setlength{\belowcaptionskip}{-0.4cm}
    \caption{Instruction examples with learned the discrete codes. In (a), the codeword responds to the real word in the LLM's vocabulary. In (b), the learned codewords are virtual tokens, which need to expand the LLM's vocabulary for fine-tuning.}
    \label{tab_instruction}
\end{table}
% \vspace{-0.2cm}

Dr.E implements token-level alignment between GNNs and LLMs.
Based on the quantized codes through intra-layer and inter-layer residuals, the preservation of multiple perspectives in each convolution step is guaranteed, fostering reliable information transfer from the surrounding nodes to the central point.
Moreover, by utilizing the LLMs’ vocabulary as the codebook, Dr.E can represent each node with real words from different graph hops, as shown in Table~\ref{tab_instruction} (a).
By fine-tuning LLaMA-2-7B~\cite{touvron2023llama2} with the designed instruction data, the node label can be easily predicted.
Different from the fixed vocabulary in Dr.E, UniMoT~\cite{DBLP:journals/corr/abs-2408-00863} expands the original LLMs' vocabulary through the incorporation of learned molecular tokens from the codebook, forming a unified molecule-text vocabulary.
Based on the specific instructions for molecular property prediction, molecule captioning, molecule-text retrieval, caption-guided molecule generation,
LLaMA2 is fine-tuned with LoRA~\cite{DBLP:conf/iclr/HuSWALWWC22},
thereby bestowing upon it an impressive and comprehensive proficiency in molecule-text applications.
% and is endowed with an impressive and comprehensive capacity for molecule-text applications.
Similarly, in KG-related scenarios, SSRQ leverages the learned discrete entity codes as additional features to directly input LLMs, helping to make accurate predictions.
Both LLaMA2-7B and LLaMA3.1-8B~\cite{DBLP:journals/corr/abs-2407-21783} are utilized for the KG link prediction task and triple classification task.
During the fine-tuning phase, the introduction of new tokens necessitates the expansion of the LLM's tokenizer vocabulary, which is in line with the size of the codebook, \eg, 1024 or 2048.
All other elements of the network remain unaltered.
Similarly, ReaLM employs LoRA fine-tuning on Llama3.2 to adapt the model for KG tasks.

\section{Future Directions}
\label{sec_future}

Based on this systematic review, we identify four key directions for future research.

\textbf{Choice of Codeword.}
For QGR learning, one can incorporate certain domain knowledge into the construction of the codebook. This knowledge might encompass semantic concepts, structural formations, and textual language. Such an approach can augment the interpretability of the whole framework, facilitating a better human understanding.

\textbf{Advanced Quantized Methods.}
QGR can leverage cutting-edge VQ methods to enhance the effectiveness and efficiency of graph codes, drawing on techniques that have already proven successful in the computer vision domain.
% QGR can adopt the latest advanced VQ methods to enhance the effectiveness and efficiency for graph codes, which may have already demonstrated success within the domain of computer vision.
For instance, the rotation trick~\cite{DBLP:journals/corr/abs-2410-06424} is a proven method to mitigate the issues of codebook collapse and underutilization in VQ.
SimVQ~\cite{zhu2024addressing} simply incorporates a linear transformation into the codebook, yielding substantial improvement.

\textbf{Unified Graph Foundation Model.}
While the QGR methods have indeed made some notable strides, the current research primarily follows a schema that tackles various tasks using distinct training targets, resulting in multiple models.
This diversity may restrict its applicability, particularly in the age of LLMs.
Drawing inspiration from the remarkable success of unified LLMs and multimodal LLMs~\cite{DBLP:journals/corr/abs-2303-18223,lin2024has}, the unified graph foundation model could be realized to execute various tasks across diverse graphs through QGR, by learning unified codes for different graphs and then tuning with LLM techniques~\cite{DBLP:conf/cikm/JiL0LL0L025}.

\textbf{Graph RAG with QGR.}
Retrieval-Augmented Generation (RAG) has emerged as a significant focal point for improving the capabilities of LLMs within specific fields.
Graph RAG~\cite{DBLP:journals/corr/abs-2404-16130,DBLP:journals/corr/abs-2503-21322} would be beneficial for downstream tasks in the context of LLMs, where the QGR learning would provide an effective manner to retrieve relevant nodes or subgraphs by calculating the similarity of the codes.
It can be easily realized by the statistical metrics of the codes or the numerical calculations among embeddings.

% \clearpage
\section*{Limitations}

Beyond the above methods, there are also some methods that do not use quantized code for the final representation, such as VQ-GNN~\cite{ding2021vq}, VQSynery~\cite{DBLP:journals/corr/abs-2403-03089}, and MSPmol~\cite{DBLP:conf/ijcnn/LuPZC24}.
For example,
instead of employing discrete motif features as the ultimate representation, MSPmol~\cite{DBLP:conf/ijcnn/LuPZC24} utilizes the Vector Quantization (VQ) method on motif nodes within specific layers of the molecular graph.
There are also benefits for the representation, which can be viewed as a paradigm of \emph{contiguous+quantized} with the network, rather than the outputs.
They differ from the studies in Table~\ref{tab_all_studies}, so we do not include them in this survey.

Despite some achievements of the QGR methods, there are still some limitations from both methodology and application perspectives~\cite{DBLP:journals/corr/abs-2507-22920}.
First, the quantization process may result in the loss of some details of the original data, which could pose challenges in areas that require accurate modeling.
Second, the quantization process requires complex techniques to ensure effective and efficient optimization, while pinepine's schema and integration with LLMs further increase the complexity of the overall framework.
Third, compared with the wide application and success of general LLMs, efforts to integrate QGR and LLMs still need to be carried out.
Leveraging multi-agent frameworks is a compelling direction~\cite{zhang2026mars,zhang2025maps}, as it could learn semantic representations to substantially refine the planning and collaborative reasoning capabilities of these systems.
% Accordingly, the direction of future development lies in the following four aspects.

\bibliography{anthology}
\bibliographystyle{acl_natbib}

\clearpage
\newpage
\newpage
% \hspace{2cm}
\appendix

\section{Merits of Quantized Graph Representation}
\label{sec_merits}

Compared to continuous representation methods, QGR offers a series of distinct advantages that fertilize its applications.

% \red{
% (1) significantly reduced memory requirements, (2) improved inference efficiency, (3) allowing Transformers to focus on long-range dependencies rather than local information, and (4) the capacity to learn more high-level representations due to a compact latent space~\cite{wang2024learning}
% }

\noindent\textbf{Embedding Parameter Efficiency.}
As large-scale graphs become ubiquitous in reality, there is a corresponding substantial increase in the representation parameters, which necessitates abundant memory usage for storage and substantial computational expenses~\cite{DBLP:conf/iclr/0001DWH22}.
Supposing $n$ as the number of nodes and $d$ as the dimensions, the required embedding parameters would be $n\times d$.
In the quantized settings, the total count of parameters would reduce to $n\times m+M\times d$, where $M$ and $m$ denote the length of the codeword set and the number of codes assigned to each node.
For example, if there are $10^6$ nodes with 1024 feature dimensions, 1024 codewords, and 8 assigned codes,
it would be a 113-fold reduction in required representation parameters.

% Parameters needed is usually large, especially as large-scale networks become more prevalent.
% \red{the trained embeddings often require a significant amount of space to store, making storage and processing a challenge, especially as large-scale networks become more prevalent.
% Although quantisation and hashing methods introduce additional information loss, they significantly reduce storage footprint and retrieval time.
% }
% $n\times d ->M\times d, n\times M, n>>M d>M$
% \red{
% $10^6$ nodes and a feature dimension of 1024 m=3, resulting in a 270-fold reduction in required memory
% }

\noindent\textbf{Explainability and Interpretability.}
The discrete nature of QGR brings the explainability and interpretability for embeddings and reasoning, as shown in Figure\ref{fig_intro} (b).
Each code would be assigned to the practical item by specific designs for direct interpretability, \eg, Dr. E~\cite{liu2024dr} introduces a language vocabulary as the code set and each node in graphs can be represented as the permutation of explainable language tokens.

\noindent\textbf{Robustness and Generalization.}
The discrete tokens of the QGR are more robust and generalizable, which usually utilizes the self-supervised strategies to learn both local-level and high-level graph structures (\eg, edge reconstruction and DGI~\cite{DBLP:conf/iclr/VelickovicFHLBH19}) as well as semantics (\eg, feature reconstruction).
% which is more robust for downstream applications.
Moreover, the acquired tokens can readily adapt to downstream applications directly or indirectly~\cite{yang2024vqgraph}.
% \red{The tokens should be robust and generalizable. To achieve this, we rely on graph self-supervised learning. Self-supervised representations have been shown to be more robust to class imbalance (Liu et al., 2022) and distribution shift (Shi et al., 2023), while also capturing better semantic information (Assran et al., 2023) compared to representations learned through supervised objectives~\cite{wang2024learning}}

\noindent\textbf{Seamless Integration with NLP Models.}
The swift advancement in natural language processing (NLP) techniques, particularly LLMs~\cite{xu2023symbol}, has sparked an increased interest in employing NLP models to resolve graph tasks.
However, the inherent representation gap between the typical graph structure and natural language poses a significant challenge to their seamless and effective integration. Graph quantization, by learning discrete codes that bear resemblance to natural language forms, can facilitate this integration directly and seamlessly~\cite{lin2024self,liu2024dr}.

\section{Algorithm for RVQ} 

To present the process more intuitively and clearly, we summarize the entire RVQ procedure in the following algorithm.
\begin{algorithm}[h]
\small
 % \DontPrintSemicolon
  \KwIn{$\widetilde{\mathbf{X}}_i$ of the encoder, codebook representation $\mathbf{C}$.
  }
  \KwOut{Quantized representation $Q(\widetilde{\mathbf{X}}_i)$.}
  Init $Q(\widetilde{\mathbf{X}}_i)=0$, $residual=\widetilde{\mathbf{X}}_i$\;
  \For{l=1 {\rm to} N$_q$ {\rm of residual iterations}}
    {  
    $Q(\widetilde{\mathbf{X}}_i)\,+\!\!=f_q(residual)$\Comment*[r]{\textrm{ Eq. (\ref{eq_vq})}}
    $residual\,-\!\!= f_q(residual)$\Comment*[r]{\textrm{ Eq. (\ref{eq_vq})}}
    }
  \textbf{Return} $Q(\widetilde{\mathbf{X}}_i)$.
 \caption{Residual Vector Quantization (RVQ).}
 \label{algorithm_rvq}
\end{algorithm}

% \newpage
\section{General Framework of QGR Studies}

The general framework of the QGR studies is illustrated in Figure~\ref{fig_arc}, which mainly comprises an encoder $f_e$, decoder $f_d$, and quantization process $f_q$.
The encoder $f_e$ is to model the original graph structure into latent space, where MLPs~\cite{he2020sneq} and GNNs~\cite{DBLP:conf/iclr/XiaZHG0LLL23} are usually utilized.
Training objectives of different levels can be utilized. 
By combining a predictor, multiple applications (including node level, edge level and graph level) can be realized.

\section{Comparison between SSQR and ReaLM}
\label{sec_ssqr_realm}

From a methodological perspective, SSQR is optimized through structure reconstruction and semantic distillation leveraging VQ techniques. Furthermore, ReaLM performs feature reconstruction using RVQ to learn discrete codes.
Specifically, SSQR calculates the quantization loss $\mathcal{L}_{q}$, the structure loss $\mathcal{L}_{st}$, and the $\mathcal{L}_{sm}$. Finally, these components are combined into a joint objective for optimization:
\begin{equation}
    \mathcal{L}_{q} = \big\| \text{sg}[\mathbf{e}^L] - \textbf{q}_e \big\|_2^2 + \beta \big\| \mathbf{e}^L - \text{sg}[\textbf{q}_e] \big\|_2^2,
\end{equation}
\begin{equation}
\label{eq_str}
  \mathcal{L}_{st}\!=\!-\frac{1}{|\mathcal{F}|}\!\sum_i [y_i\log \tilde{y}_i+(1-y_i)\log (1\!-\!\tilde{y}_i)],
\end{equation}
\begin{equation}
\label{eq_mse}
  \mathcal{L}_{se}=-\frac{1}{|\mathcal{E}|}\sum_i \big\|{\mathbf W}_{s}{\mathbf q}_{e_i}-{\mathbf t}_{e_i}\big\|_2^2,
\end{equation}
\begin{equation}
  \mathcal{L}=\mathcal{L}_{q}+\mathcal{L}_{st}+\mathcal{L}_{se}.
\end{equation}
The structural loss $\mathcal{L}_{st}$ is designed to reconstruct the KG topology (\ie, links) using the learned code representations.
Meanwhile, the semantic loss $\mathcal{L}_{sm}$ distills textual information, after embedding via LLMs, into the discrete codes.
In contrast, ReaLM first learns semantic KG embeddings using RotatE. The model then requires the quantized embedding, defined as $\mathbf{e}_q = \sum_{s=1}^{S} \mathbf{c}_{s,i_s}$, to effectively reconstruct these semantic embeddings, where $\mathbf{c}_{s,i_s}$ represents the candidate codeword representation. Incorporating the residual quantization loss, the joint optimization objective is formulated as:
\begin{equation}
    \mathcal{L}_{\text{rc}} = \underbrace{\| \mathbf{e} - \mathbf{e}_q \|^2}_{\text{reconstruction loss}} + \underbrace{\sum_{s=1}^{S} \| \text{sg}[\mathbf{res}_{s-1}] - \mathbf{c}_{s,i_s} \|^2}_{\text{codebook loss}},
\end{equation}
\begin{equation}
    \mathcal{L}_{\text{RVQ}} = \mathcal{L}_{\text{rc}} + \beta \underbrace{\sum_{s=1}^{S} \| \mathbf{res}_{s-1} - \text{sg}[\mathbf{c}_{s,i_s}] \|^2}_{\text{commitment loss}},
\end{equation}
where $\mathbf{res}_{s-1}$ is the residual vector.

To illustrate the integration process with LLMs, we present case studies for SSQR and ReaLM in Tables 3 and 4, respectively. These examples provide a comprehensive comparison of their distinct instruction formats and the resulting LLM outputs. As highlighted in \red{red} within the tables, there are two primary differences: SSQR prompts the model to generate quantized representations for the top-3 potential answer entities, whereas ReaLM only requires the top-1. Consequently, these differing instructions lead to distinct model outputs.

\section{Quantizing Graphs and Other Modalities}

Current studies primarily focus on pure graph structures or text-attributed graphs~\cite{liu2024dr,DBLP:conf/kdd/Bo0025}, which may limit their real-world applicability given the increasing importance of integrating multi-source and multi-modal information~\cite{10.1093/nsr/nwae403}.
Following this trend, MedTok~\cite{DBLP:conf/icml/0001MHJFGSZ25} jointly encodes textual descriptions from Electronic Health Records (EHRs) and KG-based representations of medical codes to generate richer and structured embeddings for medical foundation models.
Looking forward, quantizing graphs alongside other modalities~\cite{DBLP:conf/cvpr/ZhaYFRSKG25,DBLP:journals/corr/abs-2506-09110}, such as images or videos in general domains and electroencephalography (EEG) or electrocardiogram (ECG) in medical domains, could further enhance the cognitive capabilities of these models and extend their utility to a broader range of scenarios.

\section{Graph Quantization Analysis}

Although several recent approaches leverage QGR to map nodes or subgraphs to discrete tokens.
However, it remains an open question \emph{whether these quantized tokenizers effectively capture high-level, transferable graph patterns across diverse domains}.
Thus, the study~\cite{anonymous2025can} presents a comprehensive empirical study analyzing the representational consistency of quantized graph tokens across different datasets.
The Graph Token Information Discrepancy (GTID) score is introduced to quantify how well structural and feature information aligns between source and target graphs for a given token.
The findings reveal that current quantized tokenizers frequently assign identical tokens to structurally inconsistent patterns, leading to high GTID scores and diminished transfer performance.

\begin{figure*}[t]
\label{ssqr_case}
\small
\begin{tcolorbox}[title={Table 3: Case study of SSQR. Taken from~\cite{lin2024self}}]
\textbf{Input:}

This is a knowledge graph completion task, which needs to predict the tail entity for an incomplete query triplet.

The query triplet is (Valparaiso University, inverse relation of /location/location/contains, ?).

The quantized representation of entity Valparaiso University is [527] [1345] [1849] [1227] [1751] [2038] [818] [515] [1417] [333] [29] [721] [1691] [798] [1033] [153]

The answer candidates and corresponding quantized representations are as follows:

Minnesota, [1532] [258] [1837] [357] [923] [1994] [638] [555] [771] [1003] [1736] [1473] [1495] [1436] [1313] [20]

New York, [661] [1243] [542] [1741] [1907] [1799] [858] [1794] [1916] [458] [1844] [909] [438] [1737] [686] [963]

California, [1059] [1286] [1604] [846] [1086] [451] [1087] [1794] [994] [297] [1463] [159] [556] [1836] [407] [963]

Massachusetts, [202] [1243] [977] [757] [304] [389] [1172] [1308] [1916] [1858] [1323] [11] [841] [1680] [1798] [1885]

Illinois, [961] [1025] [1267] [174] [643] [1951] [1742] [1794] [1720] [1481] [543] [1883] [695] [1921] [182] [963]

New York City, [1458] [326] [1707] [239] [151] [640] [1366] [1794] [610] [458] [1844] [932] [122] [311] [121] [868]

United Kingdom, [51] [193] [1354] [669] [1867] [881] [480] [1271] [392] [1858] [650] [909] [1503] [1126] [1550] [153]

Pennsylvania, [361] [825] [1052] [1655] [1670] [732] [951] [1569] [275] [1995] [543] [4] [753] [351] [331] [637]

Los Angeles, [1584] [1231] [1707] [1461] [1867] [1466] [265] [1933] [850] [1533] [805] [1128] [1824] [1823] [307] [963]

Florida, [2016] [326] [542] [1614] [462] [1433] [1388] [819] [926] [1289] [1321] [563] [1977] [1144] [1268] [662]

Ohio, [1643] [1889] [1604] [88] [1364] [485] [1819] [1569] [54] [1582] [1500] [411] [438] [125] [1636] [20]

Texas, [2012] [1845] [1207] [412] [531] [1394] [1004] [688] [653] [1671] [1790] [1690] [1732] [1686] [1721] [1205]

Virginia, [99] [825] [738] [1859] [1287] [1540] [708] [780] [653] [662] [756] [1873] [1514] [1686] [59] [409]

England, [848] [1220] [1052] [590] [175] [451] [529] [1933] [808] [1598] [1790] [1988] [670] [524] [121] [136]

Connecticut, [633] [1025] [672] [338] [1694] [1799] [1528] [1177] [1949] [458] [1703] [411] [395] [33] [233] [1013]

Indiana, [497] [1875] [1849] [377] [1694] [61] [1471] [1445] [392] [1672] [1500] [300] [711] [1839] [331] [136]

Los Angeles County, [662] [1772] [558] [1623] [304] [1755] [1388] [1794] [1554] [333] [1662] [1258] [1474] [1764] [1021] [409]

Louisiana, [1584] [326] [1309] [50] [1035] [0] [546] [728] [1130] [297] [1284] [11] [1514] [1929] [1248] [136]

Michigan, [172] [326] [382] [174] [1364] [1774] [1165] [664] [994] [1480] [1923] [431] [816] [1587] [1443] [860]

Hennepin County, [584] [1890] [1263] [1168] [1990] [656] [784] [1735] [102] [1533] [1979] [1005] [1312] [786] [873] [1294]

Please generate quantized representations of the \red{\textbf{top-3}} potential answer entities, ranked from highest to lowest:
\\
\\
\textbf{LLM Output:}
\red{\textbf{1, [497] [1875] [1849] [377] [1694] [61] [1471] [1445] [392] [1672] [1500] [300] [711] [1839] [331] [136]
\\
2, [1532] [258] [1837] [357] [923] [1994] [638] [555] [771] [1003] [1736] [1473] [1495] [1436] [1313] [20]
\\
3, [661] [1243] [542] [1741] [1907] [1799] [858] [1794] [1916] [458] [1844] [909] [438] [1737] [686] [963]}}

\end{tcolorbox}
\end{figure*}

\begin{figure*}[t]
\label{realm_case}
\small
\begin{tcolorbox}[title={Table 4: Case study of ReaLM. Taken from~\cite{DBLP:journals/corr/abs-2510-09711}}]
\textbf{Input:}

Given the query triplet (Iron Man, has Wife, ?).

We first quantize the head entity Iron Man into a 32-entry integer code:

[7] [48] [109] [586] ... [400] [55].

The answer candidates and corresponding quantized representations are as follows:

JARVIS: [7] [51] [61] [589] ... [200] [205];

Mark VI: [7] [54] [504] [454] ... [447] [187];

Pepper Potts: [7] [67] [213] [587] ... [82] [506].

Please generate quantized representations of the \red{\textbf{top-1}} potential answer entities, ranked from highest to lowest:
\\
\\
\textbf{LLM Output:}

\red{\textbf{[6] [3] [61] [589] ... [200] [1]}}
\end{tcolorbox}
\end{figure*}

\end{document}